\documentclass[letterpaper, 10 pt, conference]{ieeeconf}  

\IEEEoverridecommandlockouts                              
\usepackage[utf8]{inputenc}
\usepackage[T1]{fontenc}

\usepackage{graphics} 
\usepackage{epsfig} 
\usepackage{mathptmx} 
\usepackage{mathptmx} 
\usepackage{amsmath} 
\usepackage{amssymb}  
\usepackage[hyperfootnotes=false]{hyperref}
\usepackage{xcolor}
\usepackage{caption}
\usepackage{booktabs}
\usepackage{tikz}
\usetikzlibrary{
    arrows.meta,
    positioning,
    fit,
    backgrounds
}

\newcommand\blfootnote[1]{%
  \begingroup
  \renewcommand\thefootnote{}%
  \footnote{#1}%
  \addtocounter{footnote}{-1}%
  \endgroup
}

\title{\LARGE \bf 
Signs of Wear: Evaluating Intelligent Speed Assistance\\
Under Synthetic Adverse Conditions
}

\author{
Johannes Fabian Hahn$^{1}$, Mao Shan$^{2}$ and Julie Stephany Berrio Perez$^{2, 3}$
\footnote{This work was supported by a scholarship of the German Academic Exchange Service (DAAD).}%
\thanks{$^{1}$Department of Informatics, University of Hamburg, Germany}
\thanks{$^{2}$Intelligent Transport Systems group at the Australian Center for Robotics, The University of Sydney, Australia. {\tt\footnotesize s.berrioperez@uq.edu.au}}
\thanks{$^{3}$School of Mechanical and Mining Engineering, University of Queensland, Australia. {\tt\footnotesize s.berrioperez@uq.edu.au}}
}

\begin{document}


\twocolumn[{%
 \renewcommand\twocolumn[1][]{#1}
 \maketitle

 \vspace{-16px}
 \begin{center}
\includegraphics[trim=0cm 0cm 0cm 0cm, clip, width=0.8\linewidth]{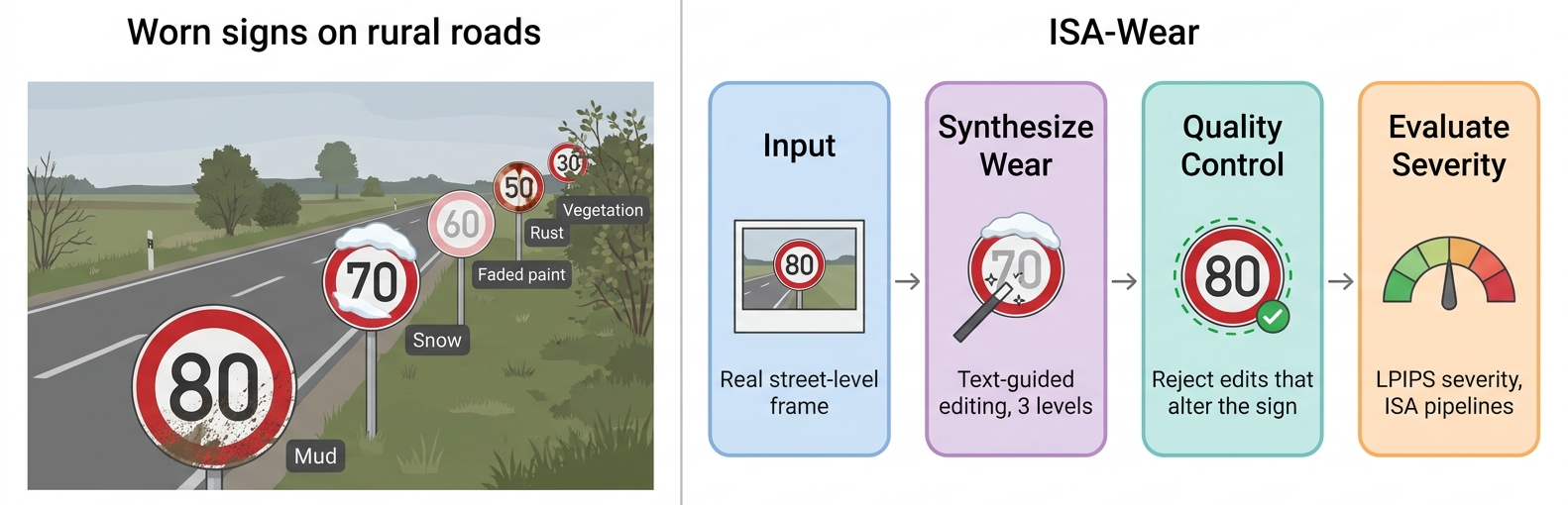}
    \captionof{figure}{\small Left: speed-limit signs on rural roads are often worn by mud, snow, faded paint, rust, or vegetation, conditions that are rare in traffic-sign benchmarks. Right: ISA-Wear adds such wear to real street-level frames with a text-guided image editor at three severity levels, rejects edits that alter the sign, and scores each edit’s severity as the LPIPS distance on the sign plate to evaluate ISA perception pipelines.}
    \label{fig:iv_teaser}
\end{center}
 \vspace{-5mm}
 }]

\thispagestyle{empty}
\pagestyle{empty}

\blfootnote{$^1$ Department of Informatics, University of Hamburg, Germany. {\tt\footnotesize johannesfabianhahn@gmail.com}}
\blfootnote{$^2$ Intelligent Transport Systems group at the Australian Center for Robotics, The University of Sydney, Australia. {\tt\footnotesize mao.shan@sydney.edu.au}}
\blfootnote{$^3$ School of Mechanical and Mining Engineering, University of Queensland, Australia. {\tt\footnotesize s.berrioperez@uq.edu.au}}

\begin{abstract}

Intelligent Speed Assistance (ISA) systems rely on cameras to recognize speed-limit signs, yet existing benchmarks underrepresent the weathered, dirty, and overgrown signs commonly encountered on rural roads. We use generative image editing to introduce different  forms of wear into real street-level images of speed-limit signs. The resulting ISA-Wear dataset contains paired original and edited images. Guided by text instructions, a multimodal image-editing model adds snow, rust, vegetation, faded paint, or mud to the speed-limit sign in a street-level frame at three severity levels. Where appropriate, the surrounding scene is adapted as well, and the viewpoint is kept. Every edited frame has an unedited counterpart, and we only evaluate frames whose original is recognized correctly, so any failure can be attributed to the wear. Since the model alters the sign's geometry in 21\% of the edits, all edits are filtered by a geometric check and a manual review. Severity is measured as the LPIPS distance between the original and the edited sign plate. On 716 worn frames, we find that in detect-then-read pipelines, wear breaks detection before it breaks reading. The detector finds only 62\% of the worn signs, even though it finds every one of them in the original frames. Missed detections account for 78--87\% of failed reads, a pattern that a zero-shot detector shares. Given the annotated sign box, as in the crop-based benchmarks, a classifier still reads 78\% of the worn signs, but only 56\% when it has to rely on the detector. Code, prompts, and the dataset are available at \texttt{[URL omitted for double-blind review]} 
\end{abstract}
\section{INTRODUCTION}
\label{sec:introduction}

Speeding is a leading contributor to fatal road crashes, and a disproportionate share of these crashes occurs on rural roads, where speeds are high, enforcement is sparse and signs are poorly maintained. Intelligent Speed Assistance (ISA) warns or limits the driver according to the posted speed limit and is among the most cost-effective countermeasures, with estimated fatal-crash reductions of 20--30\%~\cite{carsten2005isa, etsc2017isa}. ISA has been mandatory in all new vehicles in the European Union since July 2024~\cite{eu2019gsr}. The regulation allows the limit to come from a camera, from map data, or from both. In practice, the camera must cover gaps and outdated map data.

Camera-based traffic sign recognition (TSR) appears to be practically solved on standard benchmarks. Classifiers exceed human accuracy on GTSRB~\cite{stallkamp2012gtsrb} and detectors perform strongly on in-the-wild datasets such as Tsinghua--Tencent~100K and Mapillary~\cite{zhu2016tt100k, ertler2020mapillary}. However, these benchmarks mostly show well-maintained signage. On rural roads, signs fade in sunlight, rust, or become covered by snow, mud, or vegetation. These conditions are rare in training and test data, yet they are where ISA is expected to compensate for lack of enforcement. We refer to all of these as \textit{wear}: physical changes to the sign that reduce its legibility, whether lasting (faded paint, rust) or temporary (snow, mud, vegetation). Existing robustness studies apply image-level corruptions such as sensor noise, blur, or fog~\cite{hendrycks2019imagenetc, michaelis2019winter, temel2019curetsd} or add adversarial patches to the sign~\cite{eykholt2018robust}. Neither produces the realistic, natural wear of real signs nor measures its effect on ISA systems. Worn signs are too rare and too entangled with viewpoint and lighting to collect at scale.

To obtain a larger and more controlled dataset, we apply an instruction-based generative image editor to existing real-world images~\cite{rombach2022ldm, brooks2023instructpix2pix, google2025nanobanana}. The intention is to convincingly modify the sign's appearance without changing its position or the camera viewpoint. The editor takes a full street-level frame from GTSIGN-220~\cite{carnot2026gtsign} and degrades the speed-limit sign. Where it is natural, the surrounding scene is edited to match the conditions, such as snow on the ground, while viewpoint and framing stay fixed (Fig.~\ref{fig:scenes_pairs}). This gives pairs of the same image before and after weathering. Applied to the speed-limit subset of GTSIGN-220, the method yields ISA-Wear, a dataset of 883 weathered scenes under five conditions at three severity levels.

To measure the strength of each degradation, we localize the sign plate by fitting an ellipse to its red border and computing the perceptual difference between the clean and degraded signs with LPIPS~\cite{zhang2018lpips}. We measure readability with a YOLO detector followed by an OCR or a classifier reader~\cite{jocher2023ultralytics, easyocr}, which follows the detect-then-read structure of production TSR systems, and repeat the evaluation with a zero-shot detector that has never seen our training data.

Our contributions are:
\begin{itemize}
    \item a method to create paired original and worn street-level frames of speed-limit signs by generative editing, with quality control for edits that alter the sign; 
    \item a severity measure based on LPIPS computed on the localized sign plate which places all conditions on one axis; and 
    \item ISA-Wear, the resulting dataset, and an evaluation of detect-then-read ISA pipelines on it, providing insights for ADAS validation.
\end{itemize}


After reviewing related work (Sec.~\ref{sec:related}), we describe how the worn frames are generated, filtered, and scored (Sec.~\ref{sec:method}), then evaluate the pipelines on them and discuss limitations (Sec.~\ref{sec:results}), before concluding in Sec.~\ref{sec:conclusion}.
\section{Related Work}
\label{sec:related}

\textbf{Traffic Sign Recognition and ISA.} Traffic sign recognition (TSR) has been a core component of driver assistance systems for over a decade~\cite{mogelmose2012survey}. On the German Traffic Sign Recognition Benchmark (GTSRB)~\cite{stallkamp2012gtsrb}, classifiers surpassed human accuracy early on~\cite{sermanet2011multiscale, ciresan2012mcdnn}, and modern detectors achieve near-saturated performance on its detection counterpart, GTSDB~\cite{houben2013gtsdb, arcosgarcia2018evaluation}. Research has since shifted towards joint detection and classification in full street-level frames, supported by larger and more diverse datasets such as Tsinghua--Tencent~100K~\cite{zhu2016tt100k} and the Mapillary Traffic Sign Dataset~\cite{ertler2020mapillary}. These in-the-wild datasets, with many sign classes, small and occluded instances, and varied geography, remain considerably more challenging.

ISA builds directly on TSR: the system determines the current speed limit from detected signs, typically combined with map data, and informs or warns the driver when it is exceeded~\cite{eu2021isa}. Research on ISA itself has largely focused on how the system affects drivers. Field studies have found reductions in speeding and crashes \cite{vlassenroot2007belgian, carsten2005isa, lai2012benefit}, generally assuming that the speed limit displayed on a sign is correctly identified. With ISA now mandatory in the EU~\cite{eu2019gsr} and speed-assistance systems assessed in consumer safety ratings~\cite{euroncap2023sas}, the gap between benchmark accuracy and real-world perception reliability has direct safety consequences. Our work addresses this gap by evaluating speed-limit recognition under sign degradation, focusing on how wear affects detection and classification reliability in realistic driving conditions.

\textbf{Robustness of Traffic Sign Recognition.} Existing work on TSR robustness focuses mainly on adverse viewing conditions and adversarial attacks. Adverse viewing conditions include changes in lighting, weather, blur, and codec artefacts~\cite{temel2019curetsd}, synthetic image corruptions~\cite{hendrycks2019imagenetc, michaelis2019winter, kamann2020benchmarking}, rendered fog and rain~\cite{sakaridis2018foggy, tremblay2021rain}, and real-world adverse-condition datasets~\cite{sakaridis2021acdc}. In contrast, adversarial attacks modify signs with stickers or patterns designed to fool recognition models~\cite{eykholt2018robust, sitawarin2018darts}, while more recent work explores adversarial modifications that mimic natural sign damage~\cite{irshad2026advwt}. Neither directly captures naturally occurring wear, such as fading, rust, and dirt.

Although datasets of worn and damaged signs exist~\cite{floros2014degraded, chen2023mflyolo, sandhu2023condition}, evaluating the effects of wear remains challenging due to variation in viewpoint, lighting, and scene context. In this work, we focus specifically on wear applied to real images of speed-limit signs. Rather than evaluating sign crops, we retain the surrounding scene and examine how this affects both detection and speed-limit reading.

\textbf{Generative Models for Driving Perception.} Synthetic data has long addressed the cost and scarcity of real driving data, from rendered environments such as Virtual KITTI~\cite{gaidon2016virtualkitti}, SYNTHIA~\cite{RosSMVL16}, and CARLA~\cite{dosovitskiy2017carla} to simulated weather~\cite{sakaridis2018foggy, tremblay2021rain}. Diffusion models~\cite{ho2020ddpm, rombach2022ldm} now enable photorealistic image generation and editing, with synthetic images improving classifier performance~\cite{he2023synthetic, sariyildiz2023fake, azizi2023synthetic}. Instruction-based editors~\cite{brooks2023instructpix2pix} also support local image modifications, previously explored for data augmentation~\cite{trabucco2024dafusion}.

Generative models are increasingly used to evaluate robustness. ImageNet-D~\cite{zhang2024imagenet_d} generates challenging test images, while ImageNet-E~\cite{li2023imagenete} and LANCE~\cite{prabhu2023lance} edit real images to test sensitivity to changes in background, texture, and style. For traffic signs, Synset Signset Germany synthesises signs with learned wear and dirt textures~\cite{synset_signset_ger_sielemann_2024}, but does not edit real photographs. We use a commercial image-editing model~\cite{google2025nanobanana} to introduce realistic wear into real images of speed-limit signs, creating clean-degraded pairs for evaluating recognition under sign wear.

\textbf{Quantifying Image Degradation.} 
PSNR and SSIM~\cite{wang2004ssim} measure pixel-level and structural differences, but may not reflect perceptual change. LPIPS~\cite{zhang2018lpips} measures distance in pretrained network feature space, while DISTS~\cite{ding2022dists} accounts for structural and textural differences. These metrics provide alternatives for quantifying perceptual differences between images. We use LPIPS because the wear is local and structured (Sec.~\ref{sec:method:severity}).

\section{Methodology}
\label{sec:method}

\begin{table}[t]
    \centering
    \caption{\small Composition of ISA-Wear by speed limit. Clean set: of the 3{,}120 speed-limit frames in GTSIGN-220, the 1{,}000 that remain after filtering (shorter side $\geq 80$\,px, label verified) and whose annotated sign OCR reads correctly (Sec.~\ref{sec:method:data}). Sources: frames drawn from the clean set for editing. Generated: all edits, including single probe edits of sources that failed the screening. Accepted: edits that passed quality control.}
    \label{tab:dataset}
    \begin{tabular}{lrrrr}
        \toprule
        Speed limit (km/h) & Clean set & Sources & Generated & Accepted \\
        \midrule
        10 & 13 & 9 & 121 & 84 \\
        20 & 12 & 9 & 121 & 79 \\
        30 & 432 & 33 & 297 & 204 \\
        40 & 77 & 31 & 157 & 72 \\
        50 & 148 & 32 & 158 & 97 \\
        60 & 49 & 24 & 178 & 112 \\
        70 & 196 & 33 & 173 & 107 \\
        80 & 58 & 27 & 167 & 77 \\
        90 & 1 & 0 & 0 & 0 \\
        100 & 11 & 9 & 81 & 42 \\
        120 & 2 & 2 & 16 & 9 \\
        130 & 1 & 1 & 1 & 0 \\
        \midrule
        Total & 1{,}000 & 210 & 1{,}470 & 883 \\
        \bottomrule
    \end{tabular}
\end{table}

Our objective is to measure how the readability of speed-limit signs by an ISA-style perception pipeline degrades under visual deterioration typical of poorly maintained rural roads. Figure~\ref{fig:iv_teaser} summarizes the approach. Starting from real speed-limit sign images (Sec.~\ref{sec:method:data}), we synthesize degraded variants under five conditions with a generative image-editing model (Sec.~\ref{sec:method:gen}). Each original/degraded pair is then characterised by a model-agnostic severity score, obtained by fitting an ellipse to the sign’s red border and computing the perceptual distance between the two plates with LPIPS (Sec.~\ref{sec:method:severity}). Finally, both the original and the degraded images are passed through a detect-then-read ISA baseline (Sec.~\ref{sec:method:baseline}), and recognition outcomes are analysed as a function of degradation type and severity (Sec.~\ref{sec:method:protocol}).


\subsection{Source Data}
\label{sec:method:data}

We build on GTSIGN-220~\cite{carnot2026gtsign}, a dataset of 75{,}541 traffic-sign crops extracted from Mapillary street-level imagery and labelled according to the 220 sign classes of the German StVO. Because the crops originate from consumer and dash-mounted cameras rather than curated photography, they exhibit the viewpoint, resolution, and illumination variability that a vehicle-mounted camera would encounter. From the full dataset we retain only the speed-limit classes (StVO Zeichen~274), i.e.\ the circular white signs with a red border and a black numeral, covering the limits $\{10, 20, 30, 40, 50, 60, 70, 80, 90, 100, 120, 130\}$ km/h. This yields 3{,}120 images.  Since the original crops are tight around the sign, we recover the corresponding full Mapillary frames using their image IDs and retain the surrounding scene context. This allows the generative editing model to modify the complete scene rather than editing an isolated sign crop. To ensure that wear, rather than the source image, is the limiting factor for readability, we discard crops whose shorter side is below 80 pixels and a small number of samples whose labels were found to be incorrect, leaving 1{,}158. The OCR reader of Sec. \ref{sec:method:baseline} reads the correct limit on the annotated box in 1{,}000 of them (95 misread, 63 invalid); these form the clean set (Table \ref{tab:dataset}). From the clean set we draw 210 sources, grouped by speed limit.

\subsection{Generative Synthesis of Rural Degradations}
\label{sec:method:gen}

\subsubsection{Conditions}
\begin{figure}[t]
    \centering
    \includegraphics[trim={0.12cm 0 0 0}, clip, width=\linewidth]{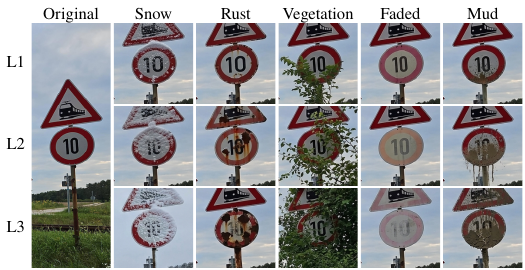}
    \caption{\small One sign (10 km/h) under five degradation conditions (columns) at three severity levels (rows, L1–L3), with the clean original on the left. All crops are taken at the same position and scale from full-frame edits.}
    \label{fig:examples_10kmh}
\end{figure}

We consider five conditions that are common on rural roads with little maintenance:
\begin{itemize}
    \item \textbf{Snow} on top of the sign and across the sign face, partly covering the numeral.
    \item \textbf{Rust}, with discoloration bleeding into the white area.
    \item \textbf{Occlusion} by branches and leaves growing in front of the sign.
    \item \textbf{Faded paint}, with the red border and the numeral losing contrast and saturation.
    \item \textbf{Mud} and road spray on the sign.
\end{itemize}
Each condition is applied at three severity levels, light (L1), moderate (L2), and heavy (L3). Fig.~\ref{fig:examples_10kmh} shows all 15 edits on one sign. We deliberately kept the signs readable even at the highest severity. Otherwise, a recognition failure could simply reflect an illegible sign rather than a limitation of the recognition pipeline.

\subsubsection{Editing Model and Prompting}
We edit with Gemini~3~Pro~Image~\cite{google2025nanobanana}, a multimodal image generation and editing model that takes a source image and a natural-language instruction and returns an edited image. Editing a real photograph, rather than generating one, maintains the viewpoint, background, and lighting of the original. The prompt describes the condition as a physical change to the sign and allows matching changes in the surroundings where they are natural, such as snow on the ground. It also states the sign's position and size, requiring both to stay fixed. Without this statement, the editor enlarged 60\% of signs in early tests. Each source is edited once per condition and level. Prompts, seeds, and the model versions are released with the dataset. The editor returns frames of about one megapixel ($1200 \times 896$ or $1376 \times 768$) with the aspect ratio of the source. The original is downscaled to the size of the edited image (area interpolation) before the plate fit, LPIPS, detection, and reading, so both frames of a pair are processed at the same resolution.

\subsubsection{Quality Control}
\begin{figure}[b]
    \centering
    \includegraphics[trim={0.9cm 0 0 0}, clip, width=\linewidth]{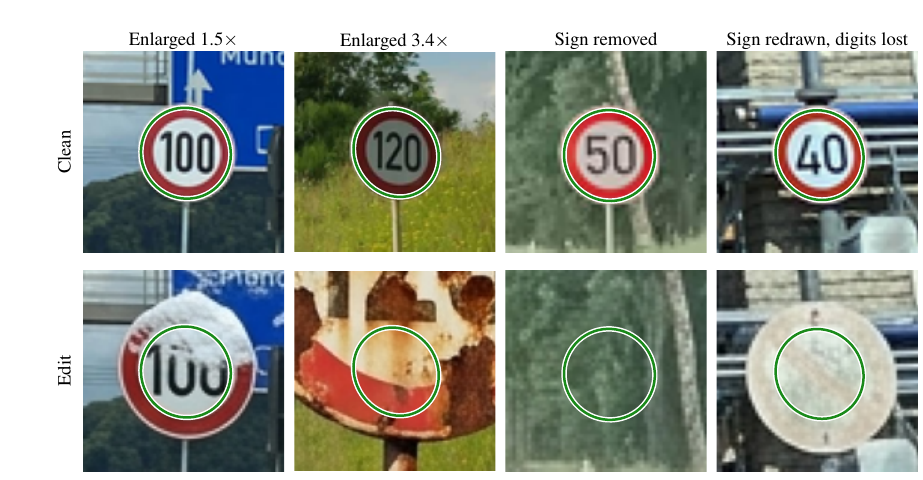}
    \caption{\small Edits rejected by quality control. Top: original; bottom: edit. The green ellipse is the original plate, drawn on both images.}
    \label{fig:failures}
\end{figure}

The editor does not always follow the instruction: it may enlarge, move, or remove the sign, or redraw it as a different one (Fig.~\ref{fig:failures}). Such samples would confound the analysis, since a recognition failure could be caused by an altered label rather than by degradation. We filter edits in two stages. Changes to the surrounding scene are intended and are not filtered.

First, a geometric consistency check compares the outer red rim of the sign in the original and the edited frame. An ellipse counts as the rim if the redness $R - max(G, B)$ just inside it exceeds the redness just outside, which separates the outer rim from the inner edge of the rim. An edit passes if the IoU of the two rims is at least $\tau_{IoU}=0.7$, which tolerates a scale change of about $1.2\times$. An edit is rejected automatically only if the rim is found in the edit and has moved or changed scale. Heavy snow, mud, or vegetation can hide the rim, so an edit whose rim cannot be verified is not rejected but passed to the manual review. Rejecting such edits would remove heavier edits because of the check itself. Positive rejections stay at 16--23\% across levels, but unverifiable rims rise from 4\% at L1 to 15\% at L3.

Second, an annotator inspects every edit that is not rejected automatically and rejects edits with a changed numeral, a replaced, added, or enlarged sign, or no visible wear. The geometric check cannot catch a changed numeral inside an intact ring, or a second sign added next to the original; the review does. In addition, every edit that a reader misreads is checked for a numeral changed by the editor, so the misread rate is not inflated by editor errors. The accept and reject criteria do not depend on any recognizer, so the filter does not favor signs that are easy to read. A reader's misreads only select edits for a second check of the numeral.

Rejected edits are not regenerated, and their rate is reported per condition and level (Sec.~\ref{sec:results}), since it shows how reliably the editor handles each condition. To limit cost, each new source first receives a single probe edit at the moderate level, with the condition rotating across sources. Only sources whose probe passes the geometric check receive the remaining 14 edits. This screens out frames on which the editor fails regardless of the condition, e.g.\ small or distant signs, and does not depend on readability. Of the 200 screened sources, 145 passed. Within the generation budget, 81 of them received further edits (79 the full set), while the probes of the other 64 are kept as single edits. Ten pilot sources received all 15 edits without screening. The 883 accepted edits therefore come from 150 sources, 89 with several edits and 61 with a single probe.

\subsection{Sign Localisation and Severity Metric}
\label{sec:method:severity}

\subsubsection{Sign Localization}

The circular plate of a speed-limit sign appears as an ellipse under camera perspective, so it can be localized geometrically, without a second learned model whose own robustness would interact with the analysis. We segment the red ring by an Otsu threshold~\cite{otsu1979threshold} on the redness map and fit an ellipse by direct least squares~\cite{fitzgibbon1999direct}. If the ring is faded or partly covered, we fit with RANSAC~\cite{fischler1981random} instead. The ellipse is fitted on the original frame only and transferred to the edited frame. After quality control the plate has the same position and scale in both, so the two crops are aligned and the score reflects a change in appearance alone.

\subsubsection{Severity Score}
The severity $d \geq 0$ of an edit is the LPIPS distance ~\cite{zhang2018lpips} between the original and the edited sign crop, computed with the AlexNet backbone and the official weights. The crop is the axis-aligned bounding square of the ellipse plus a 10\% margin, taken at the same position in both frames and resized to $128\times128$ pixels, so signs of 40 to several hundred pixels are compared at one scale. It is not rectified or masked, and so also covers the immediate surroundings of the plate. We prefer LPIPS to PSNR or SSIM~\cite{wang2004ssim} because the conditions are local and structured, where pixel metrics agree poorly with perceived change~\cite{zhang2018lpips}. LPIPS measures any visible change, not only change that matters for reading: snow on the rim and snow across the digits can give similar $d$. We therefore also compute $d_\mathrm{num}$, the LPIPS of the inner white face only, after warping the ellipse to a frontal disc at fixed resolution and masking the ring (Sec.~\ref{sec:results}). Restricting both scores to the plate excludes most changes to the scene, so they describe the condition of the sign, while the baseline still processes the full edited frame.

\begin{figure*}[t]
    \centering
    \includegraphics[width=\textwidth]{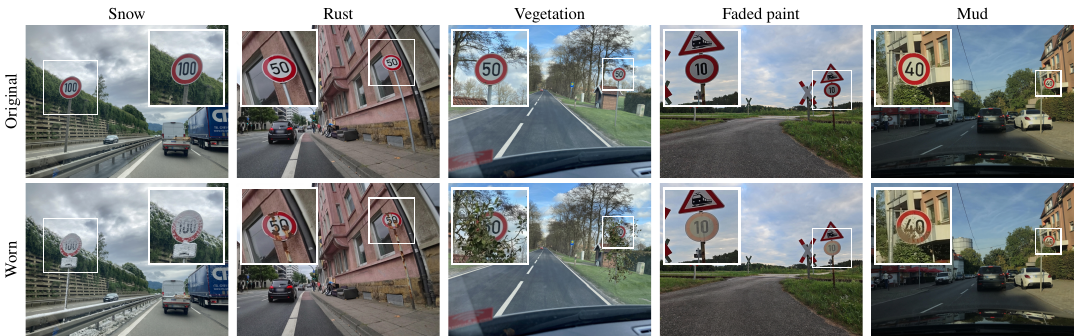}
    \caption{\small One accepted edit per condition, each in a different scene. Top: original frame; bottom: worn frame. Insets show the sign enlarged. The edit is applied to the full frame, while viewpoint and lighting are preserved.}
    \label{fig:scenes_pairs}
\end{figure*}

\subsection{ISA Perception Baseline}
\label{sec:method:baseline}

Production ISA systems are proprietary, so we build an open baseline with the same detect-then-read structure. The baseline is deliberately conventional, since our goal is to characterise the effect of degradation on a representative pipeline rather than to propose a new recogniser.

\subsubsection{Sign Detection}
A YOLOv8s~\cite{jocher2023ultralytics} detector with a single speed-limit sign class finds the sign, which leaves the classification to the reading stage. It is fine-tuned from COCO weights on the 1{,}841 GTSIGN-220 frames not used in this study for up to 100 epochs (early stopping, patience 20), with the default augmentation policy except horizontal flipping. The 1{,}000 frames of the clean set are held out from the training and validation. A sign counts as detected if a box with confidence $\geq \tau_\mathrm{det} = 0.45$ (maximum F1 on the validation split) overlaps the annotated sign with IoU $\geq$ 0.5. Otherwise the frame is a detection miss. On the 1{,}000 clean frames, the detector finds 98.3\% of the signs at 94.1\% precision.

\subsubsection{Speed-Limit Reading}
The detected box is read by two readers, one for each common TSR design. The OCR reader resizes the crop to $224 \times 224$ pixels and runs EasyOCR ~\cite{easyocr} with a digit allowlist. The classifier is a ViT-B/16~\cite{dosovitskiy2021vit}, initialized from ImageNet weights and fine-tuned for six epochs at $224\times224$ pixels on a random split of about 60{,}000 GTSIGN-220 crops of all 220 classes (97.3\% validation accuracy); a speed-limit class is mapped to its value. Both readers run on the detected box in the pipeline, and on the annotated box as an oracle that separates reading from detection. A read is \emph{correct} if it equals the true limit, a \emph{misread} if it is another legal limit in $\mathcal{S}=\{10, 20, \dots, 120, 130\}$\,km/h, and \emph{invalid} otherwise. The check against $\mathcal{S}$ mimics the plausibility check an ISA system applies before displaying a limit. A misread counts as a failure even if most digits are right, since 80 read as 30 is a safety-relevant error. Because the split is random over all crops, the classifier has probably seen crops of the source signs, so its accuracy on clean signs is optimistic. The drop under wear is less affected, since only original signs can have been seen, never their weathered versions, and every sign is read correctly in its original frame.

\subsubsection{Zero-Shot Detector}
Our detector is trained on GTSIGN-220, so its losses under wear could reflect its training data. We therefore repeat the evaluation with OWLv2~\cite{minderer2023scaling}, an open-vocabulary detector queried zero-shot with the text "speed limit sign". It has not seen GTSIGN-220; its threshold (0.43) is set at maximum F1 on 300 clean frames that are not sources of ISA-Wear, and the same two readers read its box. Since OWLv2 finds fewer signs in the original frames, all systems are compared on the frames whose original every system reads correctly (Sec.~\ref{sec:method:protocol}).

\subsection{Evaluation Protocol}
\label{sec:method:protocol}

\subsubsection{Outcomes and Metrics}
Each worn frame produces one of four outcomes for the target sign: correct, misread, invalid, or detection miss. All systems are evaluated on the same worn frames, namely those whose original every system reads correctly, so every failure can be attributed to the wear and all results share one denominator. Wear can then cause two kinds of error. Either the sign is lost, through a missed detection or an invalid read, or a wrong limit is shown, through a misread. We measure the first by detection recall, which separates failures to find the sign from failures to read it, and by readability, the share of worn frames whose sign is read correctly, i.e.\ the accuracy of the full pipeline. We measure the second by the misread rate, which plays the role of precision for ISA: a misread is how the system displays a limit that is not there, and it is reported separately because a wrong limit is more dangerous than a missing one. Detector precision in the usual sense, false detections elsewhere in the frame, is largely unaffected by the edit, which changes only the target sign, and is set by the detection threshold chosen on clean frames.

\subsubsection{Severity and Readability}

Beyond the discrete severity levels imposed by the prompts, the continuous LPIPS score enables a finer analysis. We bin samples by $d$ (and separately by $d_{\text{num}}$) and plot readability against severity for each condition, fitting a logistic model
\begin{equation}
P(\text{correct} \mid d, c) = \sigma\!\left(\alpha_c + \beta_c\, d\right)    
\end{equation}

per condition~$c$ to estimate the severity $d^{*}_c = -\alpha_c/\beta_c$ at which readability drops to 50\%. This \emph{readability threshold} places all conditions on a common perceptual scale and answers the question of how much visible deterioration a sign can sustain before an ISA system is likely to fail. To verify that LPIPS predicts readability, we report the ROC-AUC of $d$ for a failed read: the probability that a randomly chosen failed edit has a higher $d$ than a randomly chosen edit that is read correctly (0.5: chance, 1: perfect separation). The same measure compares $d$ with $d_{\text{num}}$.

\subsubsection{Implementation Details}
Edits are generated through the API of the image editor. All fine-tuning of the detector and classifier, as well as all evaluations, was performed on a single NVIDIA GTX 1080. All parameters of the plate fit, LPIPS, detector, and readers are fixed across conditions. Code, prompts, and ISA-Wear are released at \texttt{[URL omitted for double-blind review]}.

\section{Results}
\label{sec:results}

\subsection{Generation Quality}
\label{sec:results:dataset}

Of the 1{,}470 edits from 210 sources, the geometric check rejected 21\%, almost always (97\%) because the sign had been enlarged, from a subtle $1.5\times$ to a close-up of more than $3\times$ (Fig.~\ref{fig:failures}). A further 9\% had a rim that could not be verified and went to the manual review, which rejected 24\% of them. Rejection ranges from 15\% for faded paint to 65\% for rust and depends mainly on the size of the sign, not on the brightness of the image: for signs narrower than 50\,px in the edited frame (about 4\% of its width), 32\% of the edits pass, against 82\% for signs wider than 70\,px (about 5--6\%). The editor tends to redraw small signs larger, so a quality control of this kind is necessary whenever generative edits are used for evaluation. In total, 883 edits from 150 sources pass (Table~\ref{tab:dataset}). Fig.~\ref{fig:scenes_pairs} shows one accepted edit per condition in its full scene.

\subsection{Readability}
\label{sec:results:readability}
This and the following subsections use the 716 worn frames of Table~\ref{tab:results}; unless stated otherwise, the pipeline is YOLOv8s followed by either reader. Every sign is found and read correctly in its original frame, so every failure on a worn frame is caused by the wear. Wear hits the detector first: YOLOv8s misses 38\% of the worn signs, 68\% under occlusion and 58\% under snow, but only 2\% of the faded ones. Once a sign is detected, it is usually read correctly, by OCR in 83\% and by the classifier in 91\% of the cases. Missed detections therefore cause most of the failures, 78\% for OCR and 87\% for the classifier. For ISA, the type of failure matters. A missed sign is a silent failure: the system shows no limit or keeps the previous one, and the driver may not notice. An invalid read (9\% of the worn frames for OCR, 2\% for the classifier) can be recognised and discarded. A misread (2\% and 4\%) shows a wrong limit, and in 10 of the 16 OCR misreads and 14 of the 27 classifier misreads, a higher one than the true limit.

\begin{table}[t]
\centering
\caption{\small Readability (\%): share of frames in which the sign is read correctly; \emph{det.} rows: share in which it is detected. Clean: the 150 original frames. Worn conditions: the 716 worn frames whose original every system reads correctly, so all rows start from the same 100\% and share one denominator (snow 133, rust 86, occlusion 191, faded 202, mud 104 frames). All: pooled over conditions and levels L1--L3. Class.: ViT classifier.}

\label{tab:results}
\footnotesize
\setlength{\tabcolsep}{4.5pt}
\begin{tabular}{@{}llccccccc@{}}
\toprule
 & & & \multicolumn{6}{c}{Worn} \\
\cmidrule(l){4-9}
Detector & Reader & Clean & Snow & Rust & Occl. & Fade & Mud & All \\
\midrule
YOLOv8s & \textit{det.} & 97 & 42 & 57 & 32 & 98 & 77 & \textbf{62} \\
 & OCR & 95 & 35 & 55 & 19 & 87 & 59 & \textbf{51} \\
 & Class. & 97 & 33 & 53 & 26 & 93 & 71 & \textbf{56} \\
\addlinespace[3pt]
OWLv2 & \textit{det.} & 87 & 32 & 81 & 55 & 79 & 75 & \textbf{64} \\
(zero-shot) & OCR & 83 & 22 & 60 & 20 & 74 & 51 & \textbf{45} \\
 & Class. & 87 & 24 & 72 & 44 & 75 & 65 & \textbf{56} \\
\midrule
Annotated & OCR & 100 & 44 & 72 & 30 & 91 & 60 & \textbf{59} \\
box & Class. & 100 & 52 & 79 & 75 & 95 & 83 & \textbf{78} \\
\bottomrule
\end{tabular}
\end{table}

\subsection{Zero-Shot Detection}
\label{sec:results:systems}
Our detector was trained on GTSIGN-220, so its losses could reflect its training data rather than the wear. OWLv2, which has never seen these images, finds fewer of the original signs (87\% against 97\%, Table~\ref{tab:results}), but on the worn frames it loses signs at a similar rate: it detects 64\%, against 62\% for YOLOv8s. Missed detections again cause most failures, 66\% (OCR) and 82\% (classifier). The conditions differ, however: OWLv2 is less affected by rust and occlusion (81\% and 55\%, against 57\% and 32\%) and more by faded paint and snow (79\% and 32\%, against 98\% and 42\%). Detection is thus the main point of failure for both a fine-tuned and a zero-shot detector, although which wear a detector tolerates depends on the detector.

\subsection{OCR Versus Classification}
\label{sec:results:readers}
Behind YOLOv8s, the classifier stays ahead of OCR, 56\% to 51\%: it was correct where OCR failed in 52 frames, and the reverse in 17. Given the annotated box, as in an evaluation on sign crops, the gap is much larger, 78\% to 59\%. The classifier uses the whole sign, including ring and layout, and can still read signs whose digits are partly covered, but these are largely the signs the detector misses: under occlusion it reads 75\% on the annotated box and 26\% behind the detector. Better detection under wear would therefore benefit a classifier-based pipeline more than an OCR-based one.

\subsection{Severity and Readability}
\label{sec:results:severity}
We analyse severity for the YOLOv8s pipeline with both readers, on the same 716 worn frames. Readability first decreases with the prompted level. For snow, rust, and mud, OCR readability falls from 78--92\% at L1 to 15--30\% at L2 and to zero at L3. Occlusion is harmful already at L1 (38\%), whereas faded signs stay readable up to L2 and drop only at L3 (68\%). The level thus orders severity within a condition but not across conditions: a faded L3 sign is read more often than a snowy L2 sign. We therefore compare conditions on the measured severity, the plate LPIPS $d$.

\begin{figure}[t]
    \centering
    \includegraphics[width=\linewidth]{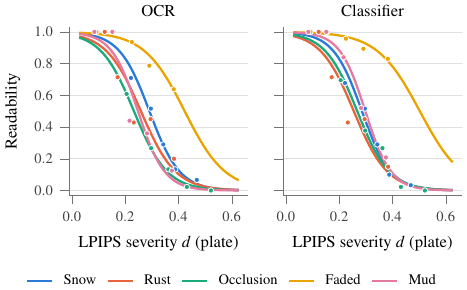}
    \caption{\small Readability of the detect-then-read pipelines against plate LPIPS. Markers: means of five equal-count bins per condition; lines: logistic fits per condition.}
    \label{fig:lpips}
\end{figure}

Readability decreases with $d$ for both readers (Fig.~\ref{fig:lpips}), and $d$ separates failed from correct reads with an ROC-AUC of 0.87 for both (0.5: no separation, 1: perfect). The editor itself causes no failures: all 45 edits whose plate barely changed ($d<0.1$) were detected and read correctly by both pipelines. Pooled over conditions, readability reaches 50\% at $d^*=0.29$ [0.27, 0.31] for OCR and $d^*=0.31$ [0.29, 0.33] for the classifier (bootstrap intervals over sources). The threshold depends on the condition: for OCR it is 0.30 for snow, 0.26 for rust, 0.23 for occlusion, and 0.25 for mud, while faded signs reach 50\% only at 0.41; in the heaviest fifth of the faded edits ($d\approx0.38$), OCR still reads 62\%, since fading lowers contrast evenly and leaves the digit shapes intact.

Restricting LPIPS to the numeral, $d_\mathrm{num}$ predicts reading on the annotated box better than $d$ (AUC 0.83 vs.\ 0.81 for OCR, 0.81 vs.\ 0.74 for the classifier) but the full pipeline worse (0.80 vs.\ 0.84 for OCR), because the detector also responds to wear on the rim and around the plate. We therefore keep $d$ as the severity of the sign.

\subsection{Limitations}
\label{sec:results:limitations}
The wear is synthetic. Whether generative wear affects recognition in the same way as real wear still has to be confirmed on real worn signs, and without a human reference we cannot say how many of the failures a driver would share. Only signs that every pipeline reads when clean are included, so the results describe how readable signs lose readability, not hard signs. We evaluate single frames, whereas production ISA tracks a sign over several frames and fuses it with map data, which can recover some missed detections. The editor fails more often on small signs, so these are under-represented. The scope is limited to German speed-limit signs in daytime Mapillary imagery, edited by one model and read by open-source baselines rather than production TSR modules. Finally, the classifier was trained on a random split of GTSIGN-220 and has probably seen the original source signs, though not their worn versions.

\section{Conclusion}
\label{sec:conclusion}
We presented a method that edits real street-level frames to show a worn speed-limit sign, and ISA-Wear, a dataset of 883 paired original and worn frames built with it. We evaluate the 716 worn frames whose original every system reads correctly, so each failure can be attributed to the edit. Severity is measured on the sign plate, independently of the system under test.

The experiments indicate that the main limitation under the tested conditions is the detection stage rather than the ability to read a sign once it has been localized. The detector found only 62\% of the worn signs, although it found all of them in the original frames, and 78--87\% of the failed reads were missed detections, while detected signs were mostly read correctly. A zero-shot detector that never saw our training data lost worn signs at a similar rate, so the effect is not an artefact of one training set. A missed sign leaves ISA without the new limit, which the driver may not notice. This failure is visible only because the wear is applied in the full frame: on the annotated box, as in an evaluation on sign crops, the classifier read 78\% of the worn signs instead of 56\%. Readability fell to 50\% at a plate LPIPS of about 0.3, which places all five conditions on one severity scale.

Camera-based ISA should therefore be tested on worn signs at controlled severity, and such tests must include detection, not only reading.

The wear is synthetic, and the editor altered the sign's geometry in 21\% of the edits and failed more often on small signs, so the dataset leans towards larger signs than typical dashcam footage. Future work should first test whether the results transfer to real wear, using naturally worn signs and a human reading reference; only then can severity thresholds inform sign maintenance. Beyond evaluation, the same method can generate worn signs for training, to test whether this makes detectors more robust to real wear. Finally, ISA combines several frames and map data, and evaluating worn signs over a drive rather than in a single frame would show how often a missed sign actually reaches the driver.

\section*{Acknowledgments}

The authors acknowledge the use of Gemini (Google), ChatGPT (OpenAI), and Claude Code (Anthropic) to assist with rephrasing, grammatical corrections, figure preparation, and software development. All AI-assisted outputs were reviewed, verified, and edited by the authors, who assume full responsibility for the final content of the manuscript.

\bibliographystyle{IEEEtran}
\bibliography{refs}

\end{document}